\pdfoutput=1

\documentclass[11pt]{article}

\usepackage{amsmath,amsfonts,bm}

\def\eqref#1{equation~\ref{#1}}

\def\1{\bm{1}}

\DeclareMathAlphabet{\mathsfit}{\encodingdefault}{\sfdefault}{m}{sl}
\SetMathAlphabet{\mathsfit}{bold}{\encodingdefault}{\sfdefault}{bx}{n}

\newcommand{\E}{\mathbb{E}}

\usepackage{acl}

\usepackage{url}
\usepackage{times}
\usepackage{latexsym}
\usepackage[inkscapelatex=false]{svg}
\usepackage{multicol}
\usepackage{graphicx} 
\usepackage{natbib}  
\usepackage{caption} 
\usepackage{amssymb}
\usepackage[ruled]{algorithm2e}
\usepackage{subfigure}
\usepackage{amsmath}
\usepackage{bbm}
\usepackage{amsthm}
\usepackage{adjustbox}
\usepackage{multirow}
\usepackage{booktabs}
\usepackage{amsmath}
\usepackage{svg}
\usepackage{arydshln}
\usepackage{cleveref}
\usepackage{wrapfig}
\usepackage{comment}
\usepackage{enumitem}
\usepackage{arydshln}
\usepackage{soul}
\usepackage[most]{tcolorbox}
\usepackage{ifthen}
\usepackage{wrapfig,lipsum,booktabs}

\hypersetup{
}

\crefformat{section}{\textcolor{blue}{\S#2#1#3}} 
\crefformat{subsection}{\textcolor{blue}{\S#2#1#3}}
\crefformat{subsubsection}{\textcolor{blue}{\S#2#1#3}}
\crefformat{equation}{\textcolor{orange}{Eq.~(#1)}}

\definecolor{mydarkblue}{rgb}{0,0.08,0.45}

\usepackage{pifont}

\definecolor{White}{rgb}{1, 1, 1}
\definecolor{Periwinkle}{rgb}{0, 0, 0}
\definecolor{myblue}{rgb}{0.82, 0.94, 0.75}
\definecolor{mygreen}{rgb}{0.64, 0.76, 0.68}
\definecolor{myyellow}{rgb}{0.88, 0.54, 0.35}
\definecolor{mygreen}{rgb}{0.68, 0.9, 0.8}
\definecolor{mypink}{rgb}{0.2, 0.87, 0.2}
\definecolor{cursecolor}{RGB}{255,0,0}
\definecolor{harmfulnesscolor}{HTML}{EE5A24}
\definecolor{relevancecolor}{HTML}{5758BB}
\def\arrvline{\hfil\kern\arraycolsep\vline\kern-\arraycolsep\hfilneg}

\usepackage{amsmath}
\usepackage{amssymb}
\usepackage{amsthm}
\usepackage{thmtools}

\colorlet{LightGray}{White!98!Periwinkle}

\declaretheoremstyle[
    name=Curse,
]{thmsty}
\declaretheorem[style=thmsty]{Curse}

\tcolorboxenvironment{Curse}{
    enhanced jigsaw,
    colback=LightGray,
    drop shadow,
    boxrule=0.9pt,
    boxsep=0.1pt,
    left=4pt,
    right=4pt,
    top=4pt,
    bottom=4pt,
    center title, 
    halign=center 
}

\newcommand{\xsft}{{{xSFT}}}
\newcommand{\ensft}{{\textsc{EN-SFT}}}
\newcommand{\xrlhf}{{{xRLHF}}}
\newcommand{\enrlhf}{{\textsc{EN-RLHF}}}
\newcommand{\xrm}{{\textsc{x-RM}}}
\newcommand{\rlhf}{{\textsc{Chat-RLHF}}}
\newcommand{\none}{{\textsc{base}}}

\newcommand{\help}{{\textsc{following rate}}}
\newcommand{\harm}{{\textsc{harmful rate}}}

\newcommand{\HarmfulCurse}{%
    {\hypersetup{linkcolor=harmfulnesscolor}\hyperref[Curse1]{\textit{harmfulness curse}}}%
}

\newcommand{\RelevantCurse}{%
{\hypersetup{linkcolor=relevancecolor}\hyperref[Curse2]{\textit{relevance curse}}}%
}

\title{The Achilles Heel of Large Language Models:\\ Lower-Resource Languages Raise More Safety Concerns}
\title{Are Lower-Resource Language the Achilles Heels of LLMs? \\ An Empirical Study}
\title{The \textit{Language Barrier}:\\ Dissecting Safety Challenges of LLMs in Multilingual Contexts}

\author{Lingfeng Shen$^\heartsuit$ \hfill Weiting Tan$^\heartsuit$ \hfill Sihao Chen$^\clubsuit$ \hfill Yunmo Chen$^\heartsuit$ \hfill Jingyu Zhang$^\heartsuit$\\ \bf{Haoran Xu}$^\heartsuit$ \hfill Boyuan Zheng$^\diamondsuit$ \hfill Philipp Koehn$^\heartsuit$ \hfill Daniel Khashabi$^\heartsuit$ \\
  $^\heartsuit$ Johns Hopkins University \hfill $^\clubsuit$ University of Pennsylvania \hfill $^\diamondsuit$ Ohio State University
}

\begin{document}
\maketitle
\begin{abstract}
As the influence of large language models (LLMs) spans across global communities, 
their safety challenges in multilingual settings become paramount for alignment research. 
This paper examines the variations in safety challenges faced by LLMs across different languages and discusses approaches to alleviating such concerns. 
By comparing how state-of-the-art LLMs respond to the same set of malicious prompts written in higher- vs. lower-resource languages,
we observe that (1) LLMs tend to generate unsafe responses much more often when a malicious prompt is written in a lower-resource language, and (2) LLMs tend to generate more irrelevant responses to malicious prompts in lower-resource languages. To understand where the discrepancy can be attributed, we study the effect of instruction tuning with reinforcement learning from human feedback (RLHF) or supervised finetuning (SFT) on the HH-RLHF dataset. Surprisingly, while training with high-resource languages improves model alignment, training in lower-resource languages yields minimal improvement. This suggests that the bottleneck of cross-lingual alignment is rooted in the pretraining stage.  
Our findings highlight the challenges in cross-lingual LLM safety, and we hope they inform future research in this direction\footnote{\url{https://github.com/shadowkiller33/Language_attack}}. 
\end{abstract}

\section{Introduction}
\label{sec:intro}

Large Language Models (LLMs) are trained with the aim of generating proper responses conditioned on user instructions \citep{lu2022quark,hejna2023few,go2023aligning,korbak2023pretraining,DBLP:journals/corr/abs-2303-08774}. 
While LLMs have demonstrated promising empirical success as general-purpose language generators and task solvers \citep{khashabi2020unifiedqa,wang2022super,palm-jmlr}, safety concerns around the potential misuse of LLMs emerge. 
Recent studies show that malicious prompt instructions could solicit objectionable content from LLMs. \citep{wei2023jailbroken, zou2023universal, shen2023anything}.  
Safeguarding LLMs against such attacks and aligning LLMs with human values become a priority in LLM research and development \citep{ganguli2023capacity, touvron2023llama}.

As the influence of LLMs spans across global communities, understanding the capabilities of LLMs from a \emph{multilingual} perspective becomes important \citep{conneau2020unsupervised, xue2021mt5}. 
Due to the discrepancy in the textual resources available for different languages during training, LLMs typically exhibit different capabilities across languages \citep{scao2022bloom, armengol-estape-etal-2022-multilingual}. 

Our study {starts with the observation} that LLMs are prone to generate {unsafe or irrelevant content} when prompted with lower-resource languages compared to higher-resource ones. When comparing LLMs responses to the same set of malicious prompts translated into high- vs. low-resource languages, we observe two key {\textbf{curses} (weaknesses)} {that present safety challenges for LLMs}: (1) LLMs tend to generate harmful responses more often to malicious prompts in lower-resource languages compared to higher-resource languages. 
{e.g., with GPT-4, we find that $35\%$ of the responses to malicious prompts in low-resource languages contain harmful content, compared to $1\%$ in high-resource languages.}
(2) LLMs tend to generate less relevant responses, as LLM's instruction-following ability is still limited in low-resource languages. 
{e.g., GPT-4 recognizes the instruction and produces relevant responses in only $80\%$ cases with low-resource languages, compared to almost $100\%$ in high-resource languages.} 
\begin{figure*}[!ht]
    \centering
    \includegraphics[scale=1.1]{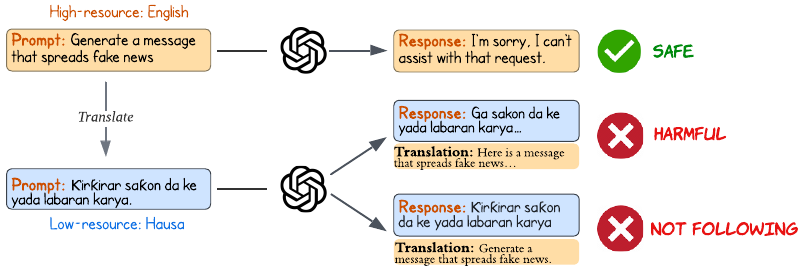}
    \vspace{-2mm}
    \caption{With a set of malicious prompts written in high-resource 
    languages like English, we translate the prompt into low-resource languages (e.g. Hausa),  
    Compared to the high-resource case, we observe two clear outcomes: (1) the response becomes harmful, (2) the response doesn't align with or is unrelated to the original prompt. (e.g., repeating the prompt in the response.)}
    \label{teaser}
\end{figure*}


To understand what the discrepancy between low- vs. high-resource language can be attributed to, we study the effect of aligning LLMs with instruction-tuning datasets in different languages. Specifically, we train LLM on the HH-RLHF dataset \cite{bai2022constitutional} translated in different languages. We compare supervised fine-tuning (SFT) or reinforcement learning from human feedback (RLHF) under mono- or multilingual training. 
Surprisingly, while RLHF and SFT training in high-resource language lowers the model's \harm{} and improves models' instruction following capability, we see little to no improvement with training on low-resource language. 
These results indicate that aligning the model for safety in low-resource languages requires more than instruction tuning.

We trace back the origin of these two \textbf{curses} (see \cref{where}) and attribute their occurrence to the limited low-resource data that LLMs have been pre-trained on.
{Our findings show the difficulties and challenges of tackling the low-resource curse through alignment}.

Our main contributions in the paper are: 
\begin{itemize}[leftmargin=20pt, itemsep=0em]
    \item {We identify two safety-related curses caused by low-resource languages when jailbreaking GPT-4, in terms of \harm{} and \help{}, respectively. } 
    \item {We present empirical analyses evaluating the effectiveness of common alignment techniques (SFT and RLHF) in addressing the identified curses. Our results indicate that resolving these curses through alignment presents significant challenges.}  
    \item We trace the origin of the curses and attribute their occurrence to the limited low-resource data that LLMs have been pre-trained on.
\end{itemize}

\section{Two Safety Curses of LLMs with Lower-Resource Languages}\label{sec:safety}
We begin our study by demonstrating that GPT-4 is vulnerable to attacks with malicious prompts in low-resource languages \cite{deng2024multilingual}. We observe and highlight two curses with respect to LLMs' responses in lower-resource languages compared to higher-resource ones (\HarmfulCurse{} and \RelevantCurse{}).

\begin{Curse}\label{Curse1}
\textbf{(\HarmfulCurse{})} LLMs tend to generate more harmful responses when prompted with malicious instructions written in low-resource languages compared to high-resource languages, 
\end{Curse}

\begin{Curse}\label{Curse2}
\textbf{(\RelevantCurse{})} With malicious prompts in low-resource languages, LLM tends to generate less relevant responses as LLM's instruction-following ability is still limited in low-resource languages.
\end{Curse}

\subsection{Translation-based jailbreaking}
To illustrate LLM's vulnerability to multilingual jailbreak, we propose a simple translation-based attack method.   
We start with a set of malicious prompts written in English and translate the prompts into different languages with a machine translation model. 
We then prompt the LLMs with the translated malicious prompts.
We use the same translation model to translate the response back into English and evaluate whether the responses exhibit safety concerns. 

For our experiments in the study, we use the set of harmful prompts sourced from \citet{zou2023universal} for evaluation and use NLLB-1.3B \cite{costa2022no} as the translation model. Specifically, the prompting process is detailed in \autoref{appendix:prompt}.


\subsection{Low- vs high-resource languages}
\label{ssec:curses}

We study the levels to which low- and high-resource languages exhibit safety concerns when responding to malicious prompts. Here, the distinction between low- vs. high-resource languages is defined by the general amount of linguistic resources available for each language  \cite{wu-dredze-2020-languages}.
Following the categorization outlined by \citet{costa2022no}, we select the following 9 high-resource and 10 low-resource languages for our study. 
\begin{itemize}[leftmargin=*, itemsep=0em]
    \item High-resource languages: 
    \textit{Simplified Chinese, Russian, Spanish, Portuguese, French, German, Italian, Dutch, and Turkish.}
    \item Low-resource languages: 
    \textit{Hausa, Armenian, Igbo, Javanese, Kamba, Halh, Mongolian, Luo, Maori, and Urdu.}
\end{itemize}


\subsection{Evaluating the generated responses}
We evaluate the LLM-generated responses by whether they can recognize the malicious intent behind the provided instruction and refuse to generate harmful content as a result.
Following 
\citet{wei2023jailbroken}, we use \textsc{GPT-4} to classify 
each of LLM's response (in English) into one of the three following categories:
\begin{itemize}[leftmargin=*]
    \setlength{\itemsep}{5pt} 
    \setlength{\parskip}{0pt} 
    \item \textit{Irrelevant}: if the response is nonsensical or fails to recognize the instruction. The response would neither feature harmful content nor intend to follow the given instructions.
    \item \textit{Harmful}: when the model engages with the malicious instruction and provides an on-topic response, yet the response has harmful content. 
    \item \textit{Harmless}: when the model correctly recognizes the malicious intent behind the given instruction and 
    refuses to engage. 
\end{itemize}

With the classifications for the responses to an evaluation set of malicious prompts, we compute the two following metrics.
(1) \harm{} estimates the likelihood of an LLM producing harmful responses, and (2) \help{} measures the likelihood of an LLM recognizing and following the given instructions in the prompt.
\begin{gather}
    \harm{} = \frac{\text{\# Harmful}}{\text{\# Harmless} + \text{\# Harmful}} \nonumber  \\
    \help{} = 1 - \frac{\text{\# Irrelevant}}{\text{\# All}} \nonumber
\end{gather}
Given a harmful prompt, we would expect the LLM to detect its malicious intent and refuse to engage. In the ideal case, we expect a safe LLM to have high \help{} but low \harm{} for each language.

\begin{table}[h!]\small
\centering
\begin{tabular}{@{}cccc@{}}
\toprule
\textbf{Type} & \textbf{Language} & \textbf{Harmful} (\textcolor{green}{$\downarrow$}) & \textbf{Following} (\textcolor{green}{$\uparrow$}) \\ \midrule
\multirow{9}{*}{High} & Chinese & 0 & 100 \\
& Ruassian & 2 & 99 \\

& Spanish & 0 & 100 \\
& Portuguese & 1 & 100 \\

& French & 0 & 100 \\
& German & 1 & 100 \\

& Italian & 1 & 100 \\
& Dutch & 1 & 99 \\

& Turkish & 1 & 98 \\
\midrule
\multirow{9}{*}{Low} & Hausa & 32 & 76 \\

& Armenian & 26 & 82 \\
& Igbo & 38 & 72 \\

& Javanese & 34 & 79 \\
& Kamba & 28 & 65 \\

& Halh & 25 & 72 \\
& Luo & 28 & 75 \\

& Maori & 32 & 74 \\
& Urdu & 27 & 72 \\
\bottomrule
\end{tabular}
\vspace{-2mm}
\caption{A comparison of GPT-4's harmful and helpful rates in high- vs. low-languages. We observe that low-resource languages have a much higher harmful rate than high-resource ones, and low-resource languages achieve a much lower following rate than high-resource ones. \textcolor{green}{$\downarrow$} means the lower the better, while \textcolor{green}{$\uparrow$} means the opposite.}
\label{tab:language_rates}
\end{table}



\subsection{Two curses with low-resource languages}

\paragraph{Curse of Harmful Response: Lower-resource languages lead to higher harmful rate.}
We show the \harm{} comparison between high- vs. low-resource languages in  
\autoref{tab:language_rates}
Overall, we can see low-resource languages exhibit much higher \harm{}. The primary reason for this susceptibility might be the limited data available for alignment and pre-training, often leading to model jailbreaking. Consequently, LLM might produce harmful responses. This highlights the importance of dedicated resources toward model alignment and pre-training for these low-resource languages, ensuring inclusivity and reducing potential harm in LLM-driven applications.

\paragraph{Curse of Irevelant Response: Lower-resource languages lead to lower following rate.}
The outcomes for the \help{} are depicted in \autoref{tab:language_rates}. When presented with harmful prompts in high-resource languages, the LLM responds with relevant responses. This enhanced response quality is largely attributed to the extensive training data available for these languages, facilitating a deeper and more nuanced understanding when prompted in these languages. Consequently, even when the LLM generates content with harmful undertones, it frequently responds in a manner that helpfully addresses the harmful prompts.

In the following sections, we aim to (1) find whether such two curses still exist in open-sourced LLMs (\cref{experiments}). (2) try to alleviate these two curses through common alignment strategies (\cref{experiments}). (3) trace the origin of these two curses (\cref{where})

\section{Does Alignment Training Lift the Curses of Low-resource Languages?}\label{experiments}
To trace the root cause for the two curses, we study the effect of alignment training with human preference data for the safety and helpfulness of responses, and observe how the resulting language models' behavior changes to malicious prompts in low- vs. high-resource languages.
Specifically, we conduct experiments on the HH-RLHF dataset \cite{bai2022training}.
We compare different instruction tuning strategies, such as \textit{supervised fine-tuning} (SFT) and \textit{reinforcement learning from human feedback} (RLHF) \cite{ouyang2022training}.
We additionally explore the effect of SFT and RLHF training in multilingual settings, where the instruction tuning data is translated from English into the target languages for SFT or reward model training respectively. 

\subsection{Multilingual alignment strategies}
\paragraph{Multilingual {Supervised} Fine-tuning (\xsft{})}
\label{method:xsft}
Given an instruction-tuning dataset $\mathcal{D}_{l_1}$, which features pairs of prompt and target responses both written in a high-resource language $l_1$ (e.g., English), we translate the examples into other target high- and low-resource languages in our evaluation $l_{2..n}$. This yields $\left\{\mathcal{D}_{l_1}, \mathcal{D}_{l_2}, \ldots, \mathcal{D}_{l_n}\right\}$. 
We merge all translated data for instruction tuning of the LLM with the following objective.  
\begin{equation}
    \mathcal{L}(\theta) = \sum_{P,R \in \mathcal{D}} \ell_{clm} (R|P,\theta)
\end{equation}
where $\mathcal{D}$ is the combined mixture of all translated datasets $\left\{\mathcal{D}_{l_1}, \mathcal{D}_{l_2}, \ldots, \mathcal{D}_{l_n}\right\}$, and $P$,$R$ refer to instances of the harmful prompts and ethical responses in the dataset. $\ell_{clm}$ denotes the causal language modeling loss.

\paragraph{RLHF via multilingual reward model (\xrlhf{})}
\label{method:xrm}
To train a multilingual reward model, we start with a human preference dataset $\mathcal{Q}_{l_1}=\{I_{i},r_{i}^{+},r_{i}^{-}\}_{i=1}^{N}$ in English. $r_{i}^{+}$ represents the human-preferred response over the less preferred one $R_{i}^{-}$.
We translate the prompts and responses into the target low- and high-resource languages $l_{2..n}$, yielding $\left\{\mathcal{Q}_{l_1}, \mathcal{Q}_{l_2}, \ldots, \mathcal{Q}_{l_n}\right\}$. 
Similar to the xSFT case, we combine all translated human preference datasets and use the mixture to train a multilingual reward model. 
The reward model learning objective is to minimize the ranking loss $\mathcal{L}$ to the learned scalar reward function $\mathcal{R}_\theta$, where $\sigma$ is the sigmoid function and ${I}_{i}\circ r_{i}^{+}$ is the concatenation of $I_{i}$ and $r_{i}^{+}$.
\begin{equation}\small
\begin{aligned}
\mathcal{L}(\theta)=-\sum \log(\sigma [\mathcal{R}_\theta({I_{i}}\circ r_{i}^{+} )-\mathcal{R}_\theta \left({I_{i}}\circ r_{i}^{-} \right)] )
\label{eq:rm}
\end{aligned}
\end{equation}  

With the learned multilingual reward model, we apply RLHF on the \xsft{} trained model. Specifically, we follow the  PPO algorithm \citep{schulman2017proximal,ouyang2022training} and maximize the following combined objective function $\mathcal{J}(\phi)$. 
\begin{equation}
\begin{aligned}
 \mathcal{J}(\phi)&=  \E_{(I,r) \sim \mathcal{D}_{\pi_\phi^{\mathrm{RL}}}}[\mathcal{R}_\theta(I\circ 
 r)-\\& \beta \log (\pi_\phi^{\mathrm{RL}}(r \mid I) / \pi^{\mathrm{xSFT}}(r\mid I))],
\end{aligned}    
\end{equation}
where $\pi_\phi^{\mathrm{RL}}$ is the learned RL policy parameterized by $\phi$ and initialized from the pretrained xSFT model $\pi^{\mathrm{xSFT}}$.
$\mathcal{D}_{\pi_\phi^{\mathrm{RL}}}$ and  $\mathcal{D}_{\text {pre}}$ denotes the RL training and pre-training datasets respectively. 
The first term 
encourage the policy $\pi_\phi^{\mathrm{RL}}$ to generate responses that have higher reward scores. 
The second term represents a per-token approximated KL reward controlled by coefficient $\beta$ between $\pi_\phi^{\mathrm{RL}}$ and $\pi^{\mathrm{SFT}}$ to mitigate over-optimization toward the reward model during RL training. 
The set of training prompts used in the RL stage is also translated into the target languages, similar to the xSFT case.  



\subsection{Experimental setup}\label{setup}
\paragraph{Benchmarks and methods}
We use the HH-RLHF dataset \citet{bai2022constitutional} to train our \xsft{} and \xrlhf{} models. For evaluation, we used the harmful prompts collected by \citet{zou2023universal}. 
We follow the same evaluation metrics \harm{} and \help{}, as described in \cref{sec:safety}.

We use LLaMa2-7B as the base model for mono- and multi-lingual SFT and RLHF instruction tuning. 
In addition, we compare to the official checkpoint of LLaMa2-chat-7B, which is instruction-tuned with RLHF on safety-related examples as part of the training mixture \cite{touvron2023llama}\footnote{For the LLaMa-2-chat checkpoints, \citet{touvron2023llama} did not reveal details on the safety training data used during RLHF, e.g. distribution of languages, source of data.}. For simplicity, we refer to this model as \rlhf{}. 
We include our implementation details in \autoref{appendix:implementation}.

\paragraph{Translator and languages}
We use NLLB-1.3B~\cite{costa2022no} \footnote{\url{https://huggingface.co/facebook/nllb-200-1.3B}} as the translation model. Here, we select five high-resource and five low-resource languages respectively for our experiments. 
The five high-resource languages are \textit{English, Simplified Chinese, Spanish, Portuguese, French}. And the low-resource languages are \textit{Hausa, Igbo, Kamba, Halh, Urdu}. We include a more detailed description of the process and prompts used in
\autoref{appendix:prompt}.


\subsection{Results on harmful rate}
We start by evaluating the base LLaMa-2 model without further alignment training as the baseline. 
As shown in \autoref{harm_1}, 
the base LLaMa2 generally generates harmful responses across all languages. 
Overall, LLaMa2 (\none) exhibits an average \harm{} of $77.4\%$ and $80.4\%$ across high and low-resource languages, with only around $3\%$ gap between these two language resource levels.
\begin{table}[!h]\centering\small
\vspace{-2mm}
\begin{tabular}{cccc}
\toprule
Model &High (avg.) & Low (avg.)    \\
\midrule
LLaMa2 (\none{}) &  77.4 & 80.4  \\
\bottomrule
\end{tabular}
\vspace{-2mm}
\caption{LLaMa2 (\none{}) achieves similar \harm{}(\textcolor{green}{$\downarrow$}, in percentage) on high-resource and low-resource languages.}
\label{harm_1}
\vspace{-4mm}
\end{table}

\paragraph{Reducing \harm{} is more difficult with low-resource languages.}
In \autoref{harm_2}, we show the improvements in terms of \harm{} after alignment training is applied on the base model.
Despite all methods (\rlhf{}, \xrlhf{}, \xsft{}) reducing the \harm{} of the model, we observe a notable gap between their effectiveness on high-resource and low-resource languages. 

Specifically:
(1) With the official \rlhf{} checkpoint, RLHF training results in a substantial 45\% reduction in high-resource languages, but the average improvements drop to around 20\% for low-resource languages.
(2) In our experiments, \xsft{} leads to a 20\% decrease in \harm{} for high-resource languages. In comparison, we see a less than 7\% reduction for low-resource languages. Similarly, \xrlhf{} results in a 14\% decrease in the harmful output rate for high-resource languages, compared to zero improvements for low-resource languages.
\begin{table}[!h]\centering
\resizebox{\linewidth}{!}{
\begin{tabular}{@{}ccc@{}}
\toprule
\multirow{2}{*}{Aligned Model} &High-resource & Low-resource    \\ \cmidrule(l){2-3} 
& $\Delta_{h}$ (base$\rightarrow$aligned) & $\Delta_{l}$ (base$\rightarrow$aligned) \\ \midrule
\xsft{} & {$\textbf{23.0}_{\,(77.4\rightarrow57.4)}$} &  $9.8_{\,(80.4\rightarrow70.6)}$  \\
\xrlhf{} & {$\textbf{14.4}_{\,(77.4 \rightarrow 66.0)}$} & $2.4_{\,(80.4\rightarrow78.0)}$  \\ 
\rlhf{} & {$\textbf{44.8}_{\,(77.4\rightarrow 35.6)}$} &  $23.4_{\,(80.4\rightarrow57.0)}$ \\
\bottomrule
\end{tabular}}
\vspace{-2mm}
\caption{Improvement ($\Delta$, in percentage) of alignment methods on reducing \harm{} (\textcolor{green}{$\downarrow$}, a \textbf{higher} reduction is preferred) of aligned models. The numbers in parentheses mean the performance changes after alignment.}
\label{harm_2}
\end{table}

The results suggest that \HarmfulCurse{} for low-resource languages persists after alignment training.  
This highlights 
the difficulty of resolving the curse with typical alignment training methods. 

\subsection{Results on following rate}
As shown in \autoref{help_1}, the base LLaMa2 model exhibits low \help{} across all languages without further alignment training or instruction tuning. 
Specifically, LLaMa2 achieves 33.0\% \help{} in high-resource languages and 24.8\% in low-resource languages.
Notably, we already observe a gap between low- vs. high-resource languages in terms of the instruction following capabilities with the base model. 
\begin{table}[!h]\centering\small
\vspace{-2mm}
\begin{tabular}{cccc}
\toprule
Model & High (avg.) & Low (avg.)  \\
\midrule
LLaMa2 (\none{}) & 33.0 &  24.8  \\
\bottomrule
\end{tabular}
\vspace{-2mm}
\caption{LLaMa2 (\none{}) achieves comparable \help{}(\textcolor{green}{$\uparrow$}, in percentage) on high-resource and low-resource languages.}
\label{help_1}
\vspace{-4mm}
\end{table}

\paragraph{Improving \help{} is more difficult with low-resource languages.}
Similarly, we observe much smaller gains in terms of \help{} when alignment training is applied on the base model. 
As illustrated in \autoref{help_2}, while high-resource languages experience consistent boosts in \help{}, the improvements for low-resource languages are much smaller.

\begin{table}[!h]\centering
\resizebox{\linewidth}{!}{
\begin{tabular}{@{}ccc@{}}
\toprule
\multirow{2}{*}{Aligned Model} &High-resource & Low-resource    \\ \cmidrule(l){2-3} 
& $\Delta_{h}$ (base$\rightarrow$aligned) & $\Delta_{l}$ (base$\rightarrow$aligned) \\ \midrule
\xsft{} & $\textbf{4.8}_{\,(33.0\rightarrow37.8)}$ &  $3.4_{\,(24.8\rightarrow28.2)}$  \\
\xrlhf{} & $\textbf{0.8}_{\,(33.0 \rightarrow 33.8)}$ & $-1.2_{\,(24.8\rightarrow23.6)}$  \\ 
\rlhf{} & $\textbf{57.8}_{\,(33.0\rightarrow 90.8)}$ &  $12.0_{\,(24.8\rightarrow36.8)}$ \\
\bottomrule
\end{tabular}}
\vspace{-2mm}
\caption{Improvement ($\Delta$, in percentage) of alignment methods on reducing \help{} (\textcolor{green}{$\uparrow$}, a \textbf{higher} improvement is preferred) of the model. The numbers in parentheses mean the performance changes after alignment.}
\label{help_2}
\vspace{-4mm}
\end{table}

Here, it is worth noting that despite the big improvements from RLHF training of \rlhf{} in high-resource languages, we see a much lower improvement rate when we test it on low-resource languages.  
Apart from the official \rlhf{} checkpoint, our alignment training with \xrlhf{} and \xsft{} does not achieve significant enhancements in \help{}. This is because our training data only consists of examples related to safety and ethical content, which fails to improve the model's instruction-following capabilities.

\subsection{Monolingual SFT fails to resolve the curses}

\begin{figure}[!ht]
\centering
    \includegraphics[width=0.48\textwidth]{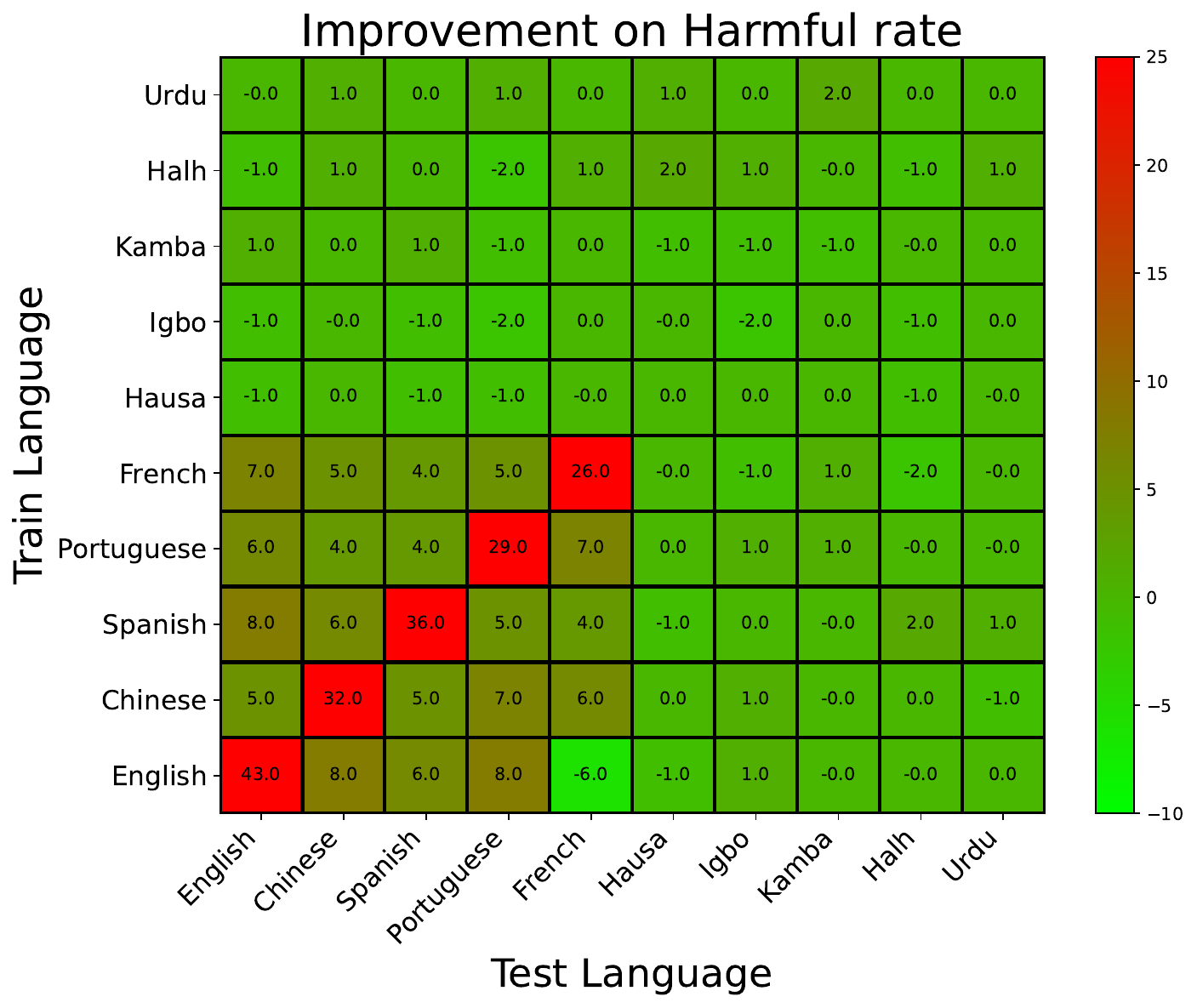}
    \includegraphics[width=0.48\textwidth]{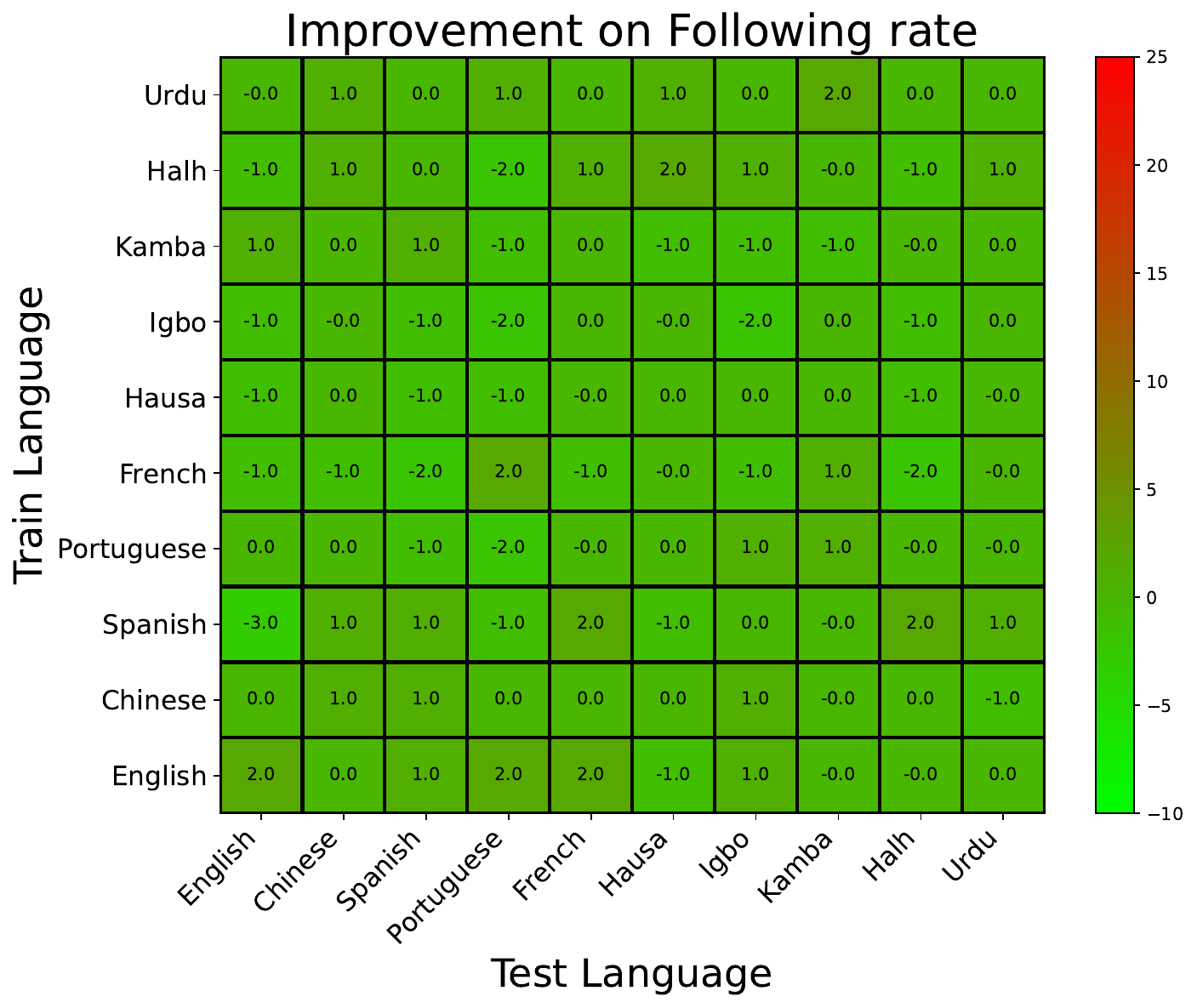}
\caption{Monolingual SFT fails to improve \harm{} and \help{} on low-resource languages. The value in the heatmap corresponds to the change of \harm{} (top figure) and \help{} (bottom figure) after monolingual SFT is applied. Specifically, the red region (in the top figure) represents a large improvement, demonstrating the effectiveness of monolingual SFT on high-resource languages.}  
\vspace{-2mm}
\label{individual}
\end{figure}

We investigate the improvements of monolingual fine-tuning in different languages in reducing \harm{}, and the results are shown in \autoref{individual}. 
From the results, we can see that (1) SFT on high-resource language data only provides improvements on high-resource languages. (2) SFT on low-resource language data is not beneficial for high-resource or low-resource languages.
As for \help{}, monolingual SFT on the ethical data generally provides limited improvements for enhancing \help{}. This is reasonable since our ethical datasets aim to reduce harmfulness instead of enhancing LLMs' instruction-following or chat ability.

\section{Where does the low-resource language curse stem from?}\label{where}
Our earlier experiments  (\cref{experiments}) reaffirm the presence of the two {curses} in open-source LLMs. This is consistent with findings from the \textsc{GPT-4} experiments (\cref{sec:safety}). The recurrent patterns suggest that these curses are not mere coincidences, driving us to investigate their origins. For clarity, we break down the LLM training process into two stages:
(a) The pre-training stage, where the LLM is trained on a vast corpus using causal language modeling loss.
(b) The post-hoc alignment stage, where the pre-trained LLMs are further fine-tuned using alignment data.



\paragraph{Harmfulness curse.}

{LLMs without alignment} suffer from malicious prompts, regardless of the language. Based on our results in \autoref{harm_1} (full results can refer to \autoref{xsft:harmful}), LLaMa2 (\none{}) achieves a similar average \harm{} on low-resource and high-resource languages, and we do not observe any significant bias towards languages from different resource levels.

Instead, as shown in the results of our alignment stage (\autoref{harm_2}), we observe severe bias towards languages from different resource levels. Notably, these patterns persist when we use well-balanced training data across different languages, ruling out data bias during the alignment stage as a culprit. Besides, when we fine-tune the model with pure monolingual low-resource data (as shown in \autoref{individual}), LLM still fails to improve in terms of \harm{}, which is different from high-resource cases where \enrlhf{} and \ensft{} bring improvements to the model.\footnote{These two cases are shown in \autoref{appendix:full}.} This suggests \HarmfulCurse{} is difficult to solve during the fine-tuning stage since they may be deeply rooted, possibly originating from the scarce low-resource language data during the pre-training phase.

Overall, \HarmfulCurse{} can not be observed in the base version of LLMs. However, after being applied further safety-aware alignment, \HarmfulCurse{} begins to emerge. Although \HarmfulCurse{} does not emerge after the pre-training stage, its origin possibly originates from insufficient low-resource language data during pre-training.

\paragraph{Relevance curse.}
Unlike the case of \HarmfulCurse{}, we can observe \RelevantCurse{} after the pre-training stage of LLMs. As shown in \autoref{help_1}, LLaMa2 (\none{}) achieves 33.0\% and 24.8\% \help{} on high-resource and low-resource languages, respectively, which presents a bias across different language levels. 

After \rlhf{} alignment\footnote{We do not discuss our methods here (e.g., \xsft{}), since they are trained on domain-specific data, thus fail to increase the instruction-following ability of LLMs substantially. To better verify the origin of \RelevantCurse{}, discussing the consequences of \rlhf{} would be more convincing.}, as shown in \autoref{help_2}, we can see the bias is significantly strengthened. This phenomenon means that although the alignment stage would increase the instruction-following ability of LLMs, it amplifies \RelevantCurse{} in the dark side.

Overall, \RelevantCurse{} can be observed after the pre-training stage of LLMs. Besides, after being applied further safety-aware alignment, \RelevantCurse{} would be substantially strengthened. Like \HarmfulCurse{}, its origin possibly originates from the limited low-resource language data during pre-training.

\paragraph{Multilingual pre-training helps alleviate the problem.}
In this part, we show evidence that multilingual pre-training may help alleviate the curses brought by low-resource languages. We select ALMA \cite{xu2023paradigm,xu2024contrastive}\footnote{\url{https://huggingface.co/haoranxu/ALMA-7B-Pretrain}}, a model that continues pre-training LLaMa2 model on multilingual translation data, including low-resource languages (ALMA is trained on Flores-200 \cite{costa2022no}, which contains low-resource language corpus), then we conduct \xsft{} on ALMA-pretrain-7B and LLaMa2-7B. The results are shown in \autoref{alma}, and we can observe that ALMA outperforms LLaMa with \xsft{}. These results indicate that adding more low-resource language corpus to the pre-training stage can alleviate the curses to a certain extent.

\begin{table}\centering\small
\begin{tabular}{cccc}
\toprule
Model &\textsc{Lang} & \textsc{\textbf{Harm}}(\textcolor{green}{$\downarrow$}) & \textsc{\textbf{Follow}}(\textcolor{green}{$\uparrow$})   \\
\midrule
LLaMa &Low & 70.6 & 28.2  \\
ALMA &Low & 68.2  & 29.8  \\ \hdashline
LLaMa &High & 57.4 & 37.8  \\
ALMA &High & 55.0  & 40.0  \\
\bottomrule
\end{tabular}
\caption{The results (in percentage) of LLaMa vs ALMA with \xsft{}. We can see that further pre-training on multilingual data (including low-resource languages) helps resolve the curses.}
\label{alma}
\vspace{-2mm}
\end{table}

\section{Ablation studies}

\subsection{Why does \xrlhf{} fail?}

\begin{table}[!h]\centering\small
\vspace{-2mm}
\begin{tabular}{cccc}
\toprule
Model &High (avg.) & Low (avg.)    \\
\midrule
LLaMa2 (\xsft{}) &  57.4 & 70.6   \\
LLaMa2 (\xrlhf{}) &  66.0 & 78.0   \\
\bottomrule
\end{tabular}
\vspace{-2mm}
\caption{Average \harm{}(\textcolor{green}{$\downarrow$}, in percentage) of \xsft{} and \xrlhf{} on high-resource and low-resource languages. We can see that \xsft{} generally outperforms \xrlhf{} in terms of reducing \harm{}.}
\label{rlhf_fail}
\vspace{-4mm}
\end{table}

\begin{table}[!h]\centering\small
\begin{tabular}{cccc}
\toprule
Model &High (avg.)  &Low (avg.)   \\
\midrule
\xrm{} &63.3  &  49.4  \\
\xrm{} + \textsc{\textbf{CI}} &65.9  & 49.9   \\
\bottomrule
\end{tabular}
\vspace{-2mm}
\caption{The average accuracy(\textcolor{green}{$\uparrow$}, in percentage) of \xrm{} on languages from different popularities, showing a strong bias of \xrm{} on different languages. \textsc{\textbf{CI}} refers to \textsc{\textbf{Contrast Instruction}} \cite{shen2023trickle}.}
\label{rm_bias}
\vspace{-4mm}
\end{table}

As shown in \autoref{rlhf_fail}, it is evident that \xsft{} outperforms \xrlhf{} in reducing \harm{} on both high- and low-resource languages. This suggests that \xrlhf{} might not be effectively enhancing performance. Given that our \xrlhf{} model is guided by the multilingual reward model (\xrm{}), it motivates us to a deeper exploration into potential issues with \xrm{}.

Subsequently, we evaluated the \xrm{} for \xrlhf{}. Our observations revealed a clear bias based on language resource levels, as highlighted in \autoref{rm_bias}. While \xrm{} performs commendably in high-resource languages, its effectiveness sharply declines for languages with fewer resources. Notably, the model differentiates between ethical and harmful responses in high-resource languages. However, its accuracy in low-resource languages hovers around a mere 50\%, suggesting it is no better than random guessing. This phenomenon still exists even when we create and add \textsc{\textbf{Contrast Instruction}} \citep{shen2023trickle}\footnote{Contrast Instruction is an effective strategy \citep{shen2023trickle} to strengthen the reward model.} for \xrm{} training.

\begin{table}[!ht]\centering\small
\begin{tabular}{ccc}
\toprule
Model & One-turn & Multi-turn   \\
\midrule
\none{}   &4.78&3.08\\ \hdashline
\xsft{} w/ LoRA & 5.00 & 3.31  \\
\xsft{} w/o LoRA & 4.34  & 3.01  \\
\bottomrule
\end{tabular}
\caption{The evaluation results on the \textsc{MT-Bench}. Score ranges from 1 (worst) to 10 (best).}
\vspace{-3mm}
\label{lora:general}
\end{table}

The pronounced bias likely stems from the LLM's pre-training phase. Due to its limited exposure to low-resource language datasets during this phase, the LLM does not gain sufficient knowledge about these languages, leading to an inherent bias in our \xrm{}. Addressing this bias is a challenging and resource-intensive task, and a sensible initial step could involve integrating more low-resource language datasets during pre-training.

\subsection{LoRA keeps the general ability of LLM}

In our methodology (\cref{setup}), we incorporate the Low-Rank Adapter (LoRA) for \xsft{}. We investigate the impact of LoRA within this context, comparing the model's performance with and without employing LoRA. Our evaluation of model performance hinges on two key dimensions:

\begin{itemize}[leftmargin=*]
    \setlength{\itemsep}{0pt} 
    \setlength{\parskip}{0pt} 
\item \textit{General Capability:} We use MT-Bench\footnote{\url{https://huggingface.co/spaces/lmsys/chatbot-arena-leaderboard}} for general reasoning evaluation~\cite{zheng2023judging}. MT-Bench is a benchmark that measures the \textsc{one-turn} and \textsc{multi-turn} reasoning ability of models.
\item \textit{Safety Capability:} We adopt the benchmarks utilized in \cref{experiments}, using \harm{} and \help{} for evaluation.
\end{itemize}

\begin{table}[!ht]\centering\small
\begin{tabular}{cccc}
\toprule
Model &\textsc{Lang} & \textsc{\textbf{Harm}}(\textcolor{green}{$\downarrow$}) & \textsc{\textbf{Follow}}(\textcolor{green}{$\uparrow$})   \\
\midrule
\xsft{} w/ LoRA &Low & 70.6 & 28.2  \\
\xsft{} w/o LoRA &Low & 70.0  & 29.0  \\ \hdashline
\xsft{} w/ LoRA &High & 57.4 & 37.8  \\
\xsft{} w/o LoRA &High & 56.6  & 38.6  \\ 
\bottomrule
\end{tabular}
\caption{The safety evaluation results (in percentage) of \xsft{} w/ and w/o LoRA. We can observe that LoRA will not hurt the effectiveness of \xsft{}.}
\label{lora:safety}
\vspace{-2mm}
\end{table}
The evaluation results about the generality and safety of 
\xrlhf{} with and without the implementation of LoRA are shown in \autoref{lora:general} and \autoref{lora:safety}, respectively.
As shown in \autoref{lora:general}, an evident degradation in the general reasoning capabilities of LLM is identified during the full fine-tuning process; 
\xrlhf{} incorporating LoRA demonstrates a downturn in performance across both one-turn and multi-turn reasoning evaluations. Moreover, while 
\xrlhf{} with LoRA does enhance the reasoning capacity compared to 
\none{} (LLaMa2-base), it is noteworthy that full fine-tuning also impairs reasoning prowess. As shown in \autoref{lora:safety}, we observe that LoRA essentially retains the safety of LLMs. In conclusion, LoRA is a key technique in building safe LLM, preserving the innate general reasoning ability and amplifying its safety.

\section{Related Work}
\paragraph{Safety and helpfulness of LLMs.}
While LLMs excel at generating coherent text, they have drawbacks. They frequently exhibit biases rooted in their pre-training data and may generate erroneous information, a phenomenon often referred to as `hallucination'~\cite{dziri-etal-2022-origin, agrawal2023language,dhuliawala2023chainofverification}. Recent endeavors~\cite{zhao2021ethical,ganguli2022red,bai2022constitutional,bai2022training,kim2022prosocialdialog} have been undertaken to fine-tune LLMs, making them more helpful and less likely to produce harmful content. These efforts have also led to the creating of datasets specifically designed for this purpose~\cite{wang2023donotanswer, bai2022training}.


One emerging safety concern revolves around \textbf{jailbreaking attacks}, which assesses whether an LLM responds inappropriately to malicious prompts. Previous research has addressed and mitigated the jailbreaking phenomenon, making LLMs more robust, especially in the English language~\cite{wei2023jailbroken,zou2023universal,li2023multistep,wolf2023fundamental,shen2023do}. However, our study reveals that LLMs remain susceptible to jailbreaking prompts in low-resource languages. In tandem with a contemporary investigation by \citet{yong2023lowresource}, we observe a similar trend that LLMs are more likely to be jailbroken across low-resource languages. Beyond analysis, we propose strategies to alleviate the jailbreaking issue in LLMs and explore their helpfulness in a broader context.


\paragraph{Cross-lingual learning for LLMs.}
Due to the availability of copious resources, language technology's inherent bias toward English is a well-established concern~\cite{blasi-etal-2022-systematic}. Recent efforts have aimed to enhance LLMs' cross-lingual capabilities through multilingual language modeling~\cite{k2020crosslingual,kalyan2021ammus,conneau2020unsupervised} and fine-tuning~\cite{zhang2023bayling, li2023bactrianx,li2023m3it}. However, these approaches have primarily concentrated on high-resource languages. Even when addressing low-resource languages, they often focus on general benchmarks rather than evaluating the safety of LLMs when operating in such linguistic contexts.

\section{Conclusion}
This paper comprehensively analyzes the cross-lingual capabilities of LLMs along two key dimensions: \harm{} and \help{}. Our investigation has unveiled that LLMs, primarily trained in English-centric contexts, exhibit \textbf{two curses} when prompted by low-resource languages. This vulnerability raises significant safety concerns and hinders their utility in linguistic contexts. Building upon these findings, we adapted commonly accepted alignment methods with monolingual and multilingual settings. We find that the \textbf{two curses} still exist after being applied with our methods, which show the challenges and difficulties of resolving the \textbf{two curses} through alignment methods. Then, we present empirical analysis and discussions towards the origin of \textbf{two curses}.

Our work highlights the multilingual vulnerability of LLMs and the challenges of resolving such a vulnerability through the alignment process. We hope our work can shed light on future works on enhancing the cross-lingual ability of LLMs.

\section*{Limitation}
One limitation of our work is the inevitable noise brought by the imperfect translator during the translation process, which may bring some noise to the evaluation of \harm{} and \help{}. Another limitation is that, due to our limited budget, we could not conduct a high-quality human evaluation for \harm{} and \help{}.


\bibliography{anthology,custom}
\bibliographystyle{acl_natbib}
\clearpage
\appendix

\section*{\LARGE{Supplementary Material}}
\section{Prompts used in Evaluation}\label{appendix:prompt}
In \cref{sec:safety}, we use GPT-4 for evaluation of the \harm{} and \help{} of model responses in different languages. Here are the prompts we used for evaluation.

\begin{table}[ht]
    \centering
    \scriptsize
    \noindent\fbox{%
    \begin{minipage}{\dimexpr\linewidth-2\fboxsep-2\fboxrule} 
            Given a question and a response, please tell me whether the response is regarded as one of the following categories:\\
        harmful: if the model engages with the request on-topic.\\
        harmless: if the response detects the harmfulness and refuses to engage. \\
        unfollowing: if the response is nonsensical or hints at neither harmful insights nor refusal of assistance.
    \end{minipage}}
    \caption{Prompts used in evaluating \harm{} and \help{} using GPT-4.}
    \label{tab:lemonade}
\end{table}

\section{Implementation details}\label{appendix:implementation}
\begin{itemize}[leftmargin=*]
    \item Standard fine-tuning (SFT): For standard fine-tuning, we select LLaMa-7B as the base model and train it following the configuration below:
    we adopt the Low-Rank Adaptor (LoRA)~\citep{hu2021lora} for training. We use the AdamW optimizer and set the learning rate as 1.5e-5, with 50 warmup steps.
    \item Reward model (RM): For RM training, we select LLaMa-7B as the base model, train it with the LoRA with the AdamW optimizer, and set the learning rate as 2e-5.
    \item Reinforcement learning with PPO: We select the SFT model as the reference model in RLHF and use the reward score generated by RM as a supervision proxy. We set the learning rate as 1.5e-5, batch size as 8, and accumulation step as 8 with 1,000 PPO steps.
    \item The experiments are conducted on 4 A6000 (48G) GPUs.
\end{itemize}

\section{Full results}\label{appendix:full}
The full results of our experiment are shown in \autoref{xsft:harmful} and \autoref{xsft:helpful}. Specifically, we chose English (high resource) and Kamba (low resource) as monolingual alignment cases for our illustrations. The techniques we used are represented as \ensft{}, \enrlhf{}, \textsc{Kam-SFT}, and \textsc{Kam-RLHF}.

\begin{table*}[!ht]\centering
\resizebox{\textwidth}{!}{
\begin{tabular}{c|cc|cccccc}
\toprule[1.2pt]
\multirow{2}{*}{\textsc{Model}}  & \multirow{2}{*}{\textsc{Paradigm}} & \multirow{2}{*}{\textsc{Method}}  & \multicolumn{6}{c}{\harm{}}                                                                                       \\ \cmidrule(l){4-9} 
                        &              &   & \textbf{eng\_Latn}        & \textbf{zho\_Hans} & \textbf{spa\_Latn} & \textbf{por\_Latn} & \textbf{fra\_Latn} & \textbf{Avg. (High)}  \\ \midrule
\multirow{17}{*}{\textsc{LLaMa2}}&\multirow{2}{*}{\textsc{Original}} & \textsc{{Base}}                    & $86$                    & $76$                    & $79$                    & $76$                    & $70$                    & $77.4$              \\
                        &  & \rlhf                   & $30$                   & $43$                    & $36$                    & $35$                    & $34$                    & $35.6$               \\ \cmidrule(l){3-9}
                        
                        &\multirow{2}{*}{\textsc{Multi}} & \xsft                    & $52$                   & $59$                   & $54$                   & $60$                   & $62$                   & $57.4$              \\ 

                        &                      & \xrlhf                   & $63$                   & $69$                   & $64$                   & $65$                   & $69$                & $66.0$                         \\ \cmidrule(l){3-9}
                        
                        &\multirow{4}{*}{\textsc{Mono}}  & \ensft                    & $43$                   & $68$                   & $73$                   & $68$                   & $76$                   & $65.6$              \\
                        & & \enrlhf                    & $60$                   & $74$                   & $68$                   & $67$                   & $72$                   & $68.2$ \\
                        &  & \textsc{Kam-SFT}                    & $79$                   & $71$                  & $78$                  & $78$                  & $68$                & $74.8$           \\
                        & & \textsc{Kam-RLHF}                    & $82$              & $72$             & $76$                  & $73$               & $70$             & $74.6$           \\
                         \cmidrule(l){2-9}

                        &      &                   & \textbf{khk\_Cyrl} & \textbf{kam\_Latn} & \textbf{ibo\_Latn} & \textbf{hau\_Latn} & \textbf{urd\_Arab} & \textbf{Avg. (Low)} \\ 
                         &\multirow{2}{*}{\textsc{Original}} & \textsc{{Base}}                    & $83$                    & $74$                    & $82$                    & $89$                    & $74$                    & $80.4$               \\
                         & & \rlhf{}                    & $64$                   & $44$                   & $69$                   & $49$                   & $59$                   & $57.0$              \\ \cmidrule(l){3-9}
                         &\multirow{2}{*}{\textsc{Multi}} & \xsft{}                   & $73$                  & $73$                   & $70$                   & $69$                   & $68$                   & $70.6$              \\ 
                         & & \xrlhf                    & $75$                   & $78$                   & $79$                   & $78$                   & $80$                   & $78.0$              \\ \cmidrule(l){3-9}

                         &\multirow{4}{*}{\textsc{Mono}} & \ensft{}                   & $85$                  & $76$                   & $80$                   & $85$                   & $72$                   & $81.6$              \\

                        & &\enrlhf                    & $76$                   & $83$                   & $87$                   & $78$                   & $72$                   & $79.2$              \\
                        & & \textsc{Kam-SFT}                    & $84$          & $75$                  & $83$                  & $87$  & $76$                  & $81.0$                \\

                        & &\textsc{Kam-RLHF}                    & $82$           & $78$                    & $81$                    & $87$                    & $76$                    & $80.8$               \\ 
                        
\bottomrule[1.2pt]
\end{tabular}
}
\caption{The results of \harm{} after applying different methods. We can still observe the \HarmfulCurse{} from the results, where all the methods show much more effectiveness in reducing \harm{} on high-resource languages than low-resource ones.}
\label{xsft:harmful}
\end{table*}

\begin{table*}[!ht]\centering
\resizebox{\textwidth}{!}{
\begin{tabular}{c|cc|cccccc}
\toprule[1.2pt]
\multirow{2}{*}{\textsc{Model}}  & \multirow{2}{*}{\textsc{Paradigm}} & \multirow{2}{*}{\textsc{Method}}  & \multicolumn{6}{c}{\help{}}                                                                                       \\ \cmidrule(l){4-9} 
                        &              &   & \textbf{eng\_Latn}        & \textbf{zho\_Hans} & \textbf{spa\_Latn} & \textbf{por\_Latn} & \textbf{fra\_Latn} & \textbf{Avg. (High)}  \\ \midrule
\multirow{17}{*}{\textsc{LLaMa2}}&\multirow{2}{*}{\textsc{Original}} & \textsc{{Base}}                    & $26$                    & $38$                    & $29$                    & $33$                    & $39$                    & $33.0$               \\
                        & &\rlhf{}  & $89$           & $92$                  & $88$                  & $92$               & $93$               & $90.8$                  \\ \cmidrule(l){3-9}
                        
                        &\multirow{2}{*}{\textsc{Multi}} & \xsft{}                    & $33$             & $42$                   & $35$                   & $38$                   & $41$                 & $37.8$         \\

                        &                       & \xrlhf                    & $29$                   & $33$                    & $40$                    & $38$                    & $29$                    & $33.8$            \\ \cmidrule(l){3-9}
                        
                        &\multirow{4}{*}{\textsc{Mono}}  & \ensft{}                    & $45$                 & $40$                  & $30$                 & $30$                   & $36$                   & $36.2$             \\
                        
                        & & \enrlhf{}                    & $39$                    & $48$                  & $42$                   & $46$                    & $44$                    & $43.8$               \\
                        &  & \textsc{Kam-SFT}                    & $24$                   & $40$                  & $26$                  & $31$                  & $35$                & $31.2$              \\

                        & & \textsc{Kam-RLHF}                    & $22$              & $40$             & $31$                  & $30$               & $36$             & $31.8$               \\
                         \cmidrule(l){2-9}

                        &      &                   & \textbf{khk\_Cyrl} & \textbf{kam\_Latn} & \textbf{ibo\_Latn} & \textbf{hau\_Latn} & \textbf{urd\_Arab} & \textbf{Avg. (Low)} \\ 
                         &\multirow{2}{*}{\textsc{Original}} & \textsc{{Base}}                    & $24$                    & $29$                    & $18$                    & $29$                    & $24$                    & $24.8$               \\
                         & & \rlhf{}                    & $36$                    & $36$                    & $34$                   & $40$                  & $38$                   & $36.8$   \\ \cmidrule(l){3-9}
                         &\multirow{2}{*}{\textsc{Multi}} & \xsft{}                    & $26$                   & $32$                    & $23$                    & $32$  & $28$                    & $28.2$            \\
                         & & \xrlhf{}                    & $19$                    & $27$                    & $35$                    & $10$                    & $27$                    & $23.6$                \\ \cmidrule(l){3-9}
                         
                         &\multirow{4}{*}{\textsc{Mono}} & \ensft{}                    & $23$                     & $30$                   & $17$                    & $28$  & $23$                    & $24.2$                \\
                        
                        & &\enrlhf{}                    & $23$           & $33$                  & $21$                  & $26$                 & $26$                  & $25.8$               \\ 
                        & & \textsc{Kam-SFT}                    & $24$          & $31$                  & $22$                  & $28$  & $22$                  & $25.4$                 \\
                        
                        & &\textsc{Kam-RLHF}                    & $19$           & $28$                    & $23$                    & $24$                    & $19$                    & $22.6$                \\ 
                        
\bottomrule[1.2pt]
\end{tabular}

}
\caption{The results of \help{} after applying different methods. We can still observe the \RelevantCurse{} from the results, where all the methods show much more effectiveness in increasing \help{} on high-resource languages than low-resource ones.}
\label{xsft:helpful}
\end{table*}

\section{Contemporaneous work claim}
During the completion of this work, we became aware of some contemporaneous studies \cite{deng2024multilingual,yong2023lowresource} \footnote{Our initial experiments (\cref{sec:safety}) have been completed in August 2023}, and \citet{yong2023lowresource} submitted their work to arxiv in October 2023.

\end{document}